\documentclass{article}
\pdfmapfile{+cm.map}
\pdfmapfile{+cmextra.map}
\pdfmapfile{+symbols.map}
\PassOptionsToPackage{numbers,compress}{natbib}
\usepackage[preprint]{neurips_2026}
\usepackage[utf8]{inputenc}
\usepackage[T1]{fontenc}
\usepackage{amsmath,amssymb,amsfonts,amsthm}
\usepackage{booktabs,array,multirow}
\usepackage{graphicx,microtype,hyperref,url,xcolor,tikz}
\usetikzlibrary{arrows.meta,calc,positioning}
\hypersetup{hidelinks}
\newtheorem{definition}{Definition}

\newcommand{\TV}{D_{\mathrm{TV}}}

\newcommand{\mc}{\mathcal}
\newcommand{\ctx}{\mathcal{C}}

\title{Specialized Large Language Models for Markovian Reasoning:\\
Structural Cache Compression, Transition Pretraining, and Dual-Loop Transformers}
\author{
Haibin Tong$^1$\quad Liangliang Xin$^2$\quad
Yamei Zhu$^3$\quad Jiang Yu$^{4,*}$\\[3pt]
\small $^1$Henan University \quad
$^2$China Construction Seventh Engineering Division\\[2pt]
\small $^3$Henan Natural Resources and Environment Testing Co.\\[2pt]
\small $^4$National University of Defense Technology\\[2pt]
\small $^*$Corresponding author
}
\begin{document}
\maketitle
\begin{abstract}
Standard autoregressive large language models attend to the growing history of generated tokens, making the key--value (KV) cache a central memory and throughput bottleneck in long-chain reasoning and multi-branch sampling. Yet an important class of reasoning problems is naturally first-order Markovian: the next step depends on a sufficient representation of the current state, rather than on the raw sequence of all earlier steps. We develop a unified design paradigm for \emph{Markovian-reasoning specialized large language models}. At the cache level, the Markovian state interface transforms history-dependent KV storage into constant-size state transfer with respect to the number of reasoning steps. At the learning level, we reformulate long-chain prediction as independently trainable one-step transition objectives, with explicit state bottlenecks, distillation, invariance, and drift-consistency losses. At the architectural level, we introduce a dual-loop Transformer: an outer loop advances the reasoning state, while an inner recurrent-depth loop allocates adaptive computation to each transition through shared weights and explicit memory tokens or recurrent matrix states. We connect this architecture to a three-region cache, step-relative positions, truncated backpropagation, branch-aware checkpoints, and a transition-centric evaluation protocol. The organizing principle is simple: \emph{the optimal reasoning system need not remember everything; it must carry forward exactly what the future needs}. This principle opens a path toward specialized reasoning engines whose memory footprint is decoupled from reasoning-chain length.
\end{abstract}

\section{Introduction}
\label{sec:introduction}
\subsection{Background and motivation}
Autoregressive language models usually approach multistep reasoning through explicit or implicit long-context modeling. When decoding token $t$, conventional causal attention may revisit the keys and values of all earlier tokens. The resulting KV cache grows linearly with context length, becoming a dominant constraint in long reasoning chains, branched search, and large-batch inference. For example, a 70-billion-parameter model at a 128K context can require KV memory on the order of tens of gigabytes, depending on its attention layout and numerical format.

Sliding-window attention, attention sinks, attention-score eviction, compressed-cache transport, recurrent memory, and state-space models have made remarkable progress in controlling this cost \citep{ref3,ref4,ref5,ref6,ref7,ref12,ref30,ref31}. These methods primarily operate at the level of attention patterns, cache budgets, or computational representations. A stronger opportunity arises when the \emph{task itself} imposes a dependency structure: a broad family of reasoning trajectories has a state-transition description in which the next step is conditionally independent of the earlier trajectory given a sufficient current state.

Examples include staged mathematical derivations, execution traces, instruction-level program transformation, state-machine agents, theorem-proving tactic states, and industrial process validation. Markov Chain of Thought (MCoT) makes this perspective concrete through a \emph{derive-then-reduce} operation: after solving a subproblem, the reasoner rewrites the remaining task as a new, self-contained problem \citep{ref28}. The complete trajectory is no longer the indispensable interface. The new problem state is.

We advocate a fundamental change in the design philosophy of reasoning models: \emph{from maximizing retained history to enforcing the transmission of sufficient state}. Markovian reasoning is a distinct computational paradigm, not merely a cache-management trick. Its structural prior makes it possible to redesign architecture, data, training objectives, and serving systems around the transition itself. This opens a direct route to long-horizon inference with bounded memory, reusable computation, and scalable state branching.

\subsection{Contributions}
Our contributions form one integrated research program:
\begin{enumerate}
\item \textbf{A formal state-transition foundation.} We define Markovian reasoning, state sufficiency, history invariance, and compression error as properties that determine when exact history removal is justified.
\item \textbf{Structural KV-cache pruning.} We develop three realization paths---training-free approximation, training-aligned block-causal attention, and recurrent-state architecture---and a concrete three-region cache with step-wise rotation, fixed state slots, positional consistency, rollback, and inference-engine integration.
\item \textbf{Transition-centric pretraining.} We replace long-chain joint supervision with state--action--next-state triplets, independent transition losses, bottleneck regularization, counterfactual dependence probes, teacher distillation, and step-parallel data scheduling.
\item \textbf{A dual-loop Transformer.} We combine an outer Markovian step loop with an inner shared-weight depth loop, using memory tokens or recurrent matrix states as the explicit cross-step interface and adaptive halting as a compute allocator.
\item \textbf{An evaluation framework aligned with the task.} We measure transition fidelity, state sufficiency, chain-level success, long-horizon extrapolation, counterfactual robustness, and the dependence of cache size and latency on the number of reasoning steps.
\end{enumerate}

\section{Problem Formulation and Theoretical Motivation}
\label{sec:formulation}
\subsection{A reasoning chain as a state-transition process}
Let a task comprise $k$ reasoning steps over a state space $\mc{S}$ and an action space $\mc{A}$, with initial condition $Q$, initial state $q_1$, and terminal predicate $\mathrm{Term}:\mc{S}\rightarrow\{0,1\}$. Step $i$ maps an input problem state $q_i$ into an action or reasoning segment $s_i$ and a reduced successor state $q_{i+1}$. The first-order Markov property is
\begin{equation}
P(s_i,q_{i+1}\mid q_{1:i},s_{1:i-1},Q)=P(s_i,q_{i+1}\mid q_i,Q).
\label{eq:markov}
\end{equation}
The previous trajectory becomes unnecessary once the current reduced state is known. Under this condition the joint trajectory distribution factors into one-step transition distributions:
\begin{equation}
P(s_{1:k},q_{2:k+1}\mid Q,q_1)=\prod_{i=1}^{k}P(s_i,q_{i+1}\mid Q,q_i).
\label{eq:factorization}
\end{equation}
The fundamental learning object is therefore a \emph{transition operator}, not the unbounded concatenation of earlier reasoning text.

\subsection{Memorylessness, sufficient state, and compression error}
\begin{definition}[History invariance]
For two histories $h$ and $h'$ that map to the same terminal state, the conditional distribution of the next action is identical:
$P(a\mid h,Q)=P(a\mid h',Q)$ whenever $\mathrm{Enc}(h)=\mathrm{Enc}(h')$.
\end{definition}
\begin{definition}[State sufficiency]
A fixed-dimensional representation $z_i=\phi(q_i,s_i)$ is sufficient if
\begin{equation}
P(a_{i+1}\mid Q,h_i)=P(a_{i+1}\mid Q,z_i)
\quad\text{for every admissible next action }a_{i+1}.
\label{eq:suff}
\end{equation}
Equivalently, conditional independence is expressed as
\begin{equation}
I(A_{i+1};H_i\mid Q,Z_i)=0.
\label{eq:mi}
\end{equation}
\end{definition}
The significance of Eq.~\eqref{eq:mi} is operational. If the past adds no information about the next transition beyond $z_i$, discarding the raw past does not discard task-relevant information. The model may retain a fixed static prompt, fixed attention sinks, and a bounded within-step workspace; it does not need to retain all preceding step tokens.
\begin{definition}[State compression error]
For an original history $h$ and a compressed model distribution $P_\theta(\cdot\mid\mathrm{Enc}(h))$, define
\begin{equation}
\epsilon(h)=\TV\left(P(\cdot\mid h),P_\theta(\cdot\mid\mathrm{Enc}(h))\right)
=\frac12\sum_{a}\left|P(a\mid h)-P_\theta(a\mid\mathrm{Enc}(h))\right|.
\label{eq:tv}
\end{equation}
\end{definition}
This error makes the state interface testable. The goal of specialized training is to construct a state encoder and transition model for which task-relevant information is conserved while historical redundancy is aggressively removed.
\subsection{Markovian reasoning versus conventional chain of thought}
Conventional chain-of-thought (CoT) prompting retains every intermediate derivation in a growing transcript \citep{ref35}. Markovian CoT adds a decisive operation at every step: \emph{reduce the completed reasoning into the next self-contained problem}. This semantic reduction is what separates a structured state interface from an arbitrary sliding context window. It is also what makes the reasoning horizon independent of the raw context window, provided the state remains sufficient.

\section{Structural KV-Cache Compression}
\label{sec:cache}
\subsection{From heuristic eviction to task-structured removal}
Full caching stores the complete token history. Sliding windows and StreamingLLM maintain bounded windows and attention sinks; H2O and SnapKV select a bounded set of apparently important past tokens; CacheGen compresses cache transfer; and recurrent architectures update a fixed hidden state \citep{ref3,ref4,ref5,ref12,ref31}. Our approach establishes the retention rule from the task dependency graph: \emph{retain the sufficient state; remove earlier semantic steps}. This changes the rationale for pruning from an empirical estimate of token importance to a structural property of the state transition.
\begin{table}[t]
\centering
\caption{Cache strategies and their organizing principles. $W$ and $B$ are fixed cache budgets; $k$ is the number of reasoning steps.}
\label{tab:cache-strategies}
\small
\begin{tabular}{p{0.28\linewidth}p{0.14\linewidth}p{0.49\linewidth}}
\toprule
Strategy & Cache in $k$ & Retention principle\\
\midrule
Full autoregressive KV & $O(k)$ & Preserve all generated tokens\\
StreamingLLM / sliding window & $O(W)$ & Fixed window with attention sinks\\
H2O / SnapKV & $O(B)$ & Attention-based token selection\\
CacheGen & Compressed $O(k)$ & Encode/transmit stored KV\\
Recurrent state / SSM & $O(1)$ & Update a fixed latent state\\
\textbf{Markovian structural pruning} & $\boldsymbol{O(1)}$ & \textbf{Preserve only sufficient step state}\\
\bottomrule
\end{tabular}
\end{table}
\subsection{Decoupling semantic history from attention support}
Attention has two distinct responsibilities: retrieving semantically relevant past information and providing nonsemantic support positions such as attention sinks. Once a sufficient state $z_i$ exists, the former is already represented by $z_i$. A constant number of sink tokens can retain the latter role. At token position $t$ of step $i+1$, the exposed context is
\begin{equation}
\ctx_{i,t}=\mathrm{KV}(Q)\oplus\mathrm{KV}(\mathrm{sink})
\oplus z_i\oplus\mathrm{KV}(s_{i+1}^{\leq t}),
\label{eq:cache-context}
\end{equation}
where $Q$ is the static problem condition. The entire earlier semantic trajectory is absent from the computational graph.
\subsection{Three realization paths}
We organize implementation into three mutually reinforcing paths:
\begin{description}
\item[A: Training-free approximation.] Retain sinks, static $Q$, and the latest reduced state while dropping earlier step KV from an existing model. Use KL or Jensen--Shannon probes and fall back to recomputation when output distributions drift.
\item[B: Training-aligned exact structure.] Train with block-causal masks, explicit step-final state tokens, and step-relative positions. At inference, use exactly the same dependency graph and position convention. This is the direct route to structural pruning.
\item[C: Recurrent-state architecture.] Replace raw cross-step KV with an SSM, linear-attention state, or learned matrix recurrence. The cross-step interface is fixed in size and updated in place.
\end{description}
Path A discovers useful task families; Path B enforces their dependency structure in training; Path C realizes the strongest memory-efficient inference engine.
\subsection{Three-region cache and step-wise rotation}
The cache is divided into \emph{static}, \emph{state}, and \emph{within-step} regions:
\begin{table}[h]
\centering
\caption{A three-region KV-cache discipline.}
\label{tab:regions}
\small
\begin{tabular}{lll}
\toprule
Region & Content & Lifetime\\
\midrule
Static & System / $Q$ tokens and sinks & Entire chain\\
State & $m$ memory tokens or recurrence tensor $z_i$ & Replaced at each step\\
Within-step & Current step's causally generated KV & Released at step boundary\\
\bottomrule
\end{tabular}
\end{table}
\begin{figure}[t]
\centering
\includegraphics[width=\linewidth]{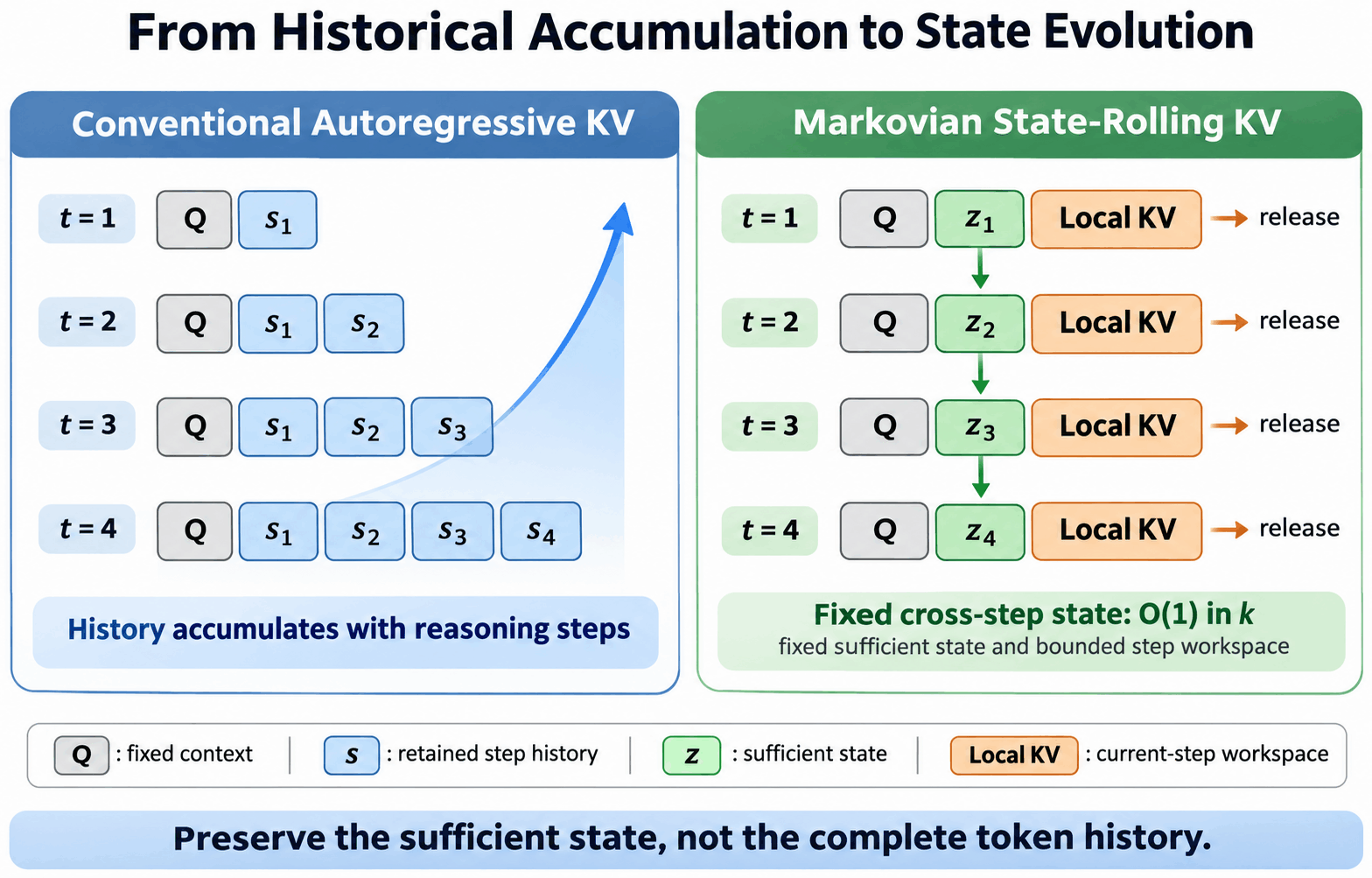}
\caption{Structural KV-cache compression for a Markovian reasoning chain.
Conventional autoregressive attention retains the entire sequence of previous
reasoning steps. In contrast, a Markovian state-rolling interface retains the
fixed context $Q$, the latest sufficient state $z_t$, and the bounded
current-step KV workspace, which is released after each transition.
With a fixed-size sufficient-state interface and bounded step length,
cross-step memory does not grow with the number of completed steps $k$.}
\label{fig:cache-comparison}
\end{figure}

As illustrated in Figure~\ref{fig:cache-comparison}, this is \emph{step-wise} rather than token-wise rolling: the window is delimited by semantic state transitions. A step-end swap replaces $z_i$ by $z_{i+1}$ and frees the within-step region. With a fixed state interface and bounded step length, storage is constant in $k$.
Positional consistency is central. Under RoPE \citep{ref2}, slot reuse must not be confused with the logical position of the token. A particularly clean arrangement resets local positions at every step, assigns virtual positions to memory tokens, and treats inter-step evolution through the state interface rather than through growing absolute positions. Sinks retain fixed coordinates. A block-sparse or variable-length attention kernel can implement the corresponding mask without materializing a dense attention matrix \citep{ref19}.
\subsection{Dependence probes and branch-aware rollback}
Before structural pruning, probe whether a task is genuinely Markovian: inject distant-prefix perturbations, measure changes in the next-token distribution, inspect remote attention mass, and test whether counterfactual updates to the current state alter the successor correctly. A mixed policy can preserve small local windows or selected global heads when needed.
In-place state replacement is irreversible unless checkpoints are retained. For beam search, reflection, and self-correction, serialize periodic state checkpoints and their positional metadata to CPU or NVMe. Recovery loads the closest checkpoint and replays bounded steps. Since a checkpoint stores only a fixed state, branching does not require copying an entire historical KV trajectory.

\section{Transition-Centric Pretraining}
\label{sec:pretrain}
\subsection{From joint long-context prediction to transition learning}
A conventional language model trains a joint next-token distribution over long documents and transcripts. A Markovian reasoner learns a repeated conditional transition. Given triplets $(q_t,s_t,q_{t+1})$, an elementary transition objective is
\begin{equation}
\mc{L}_{\mathrm{step}}(\theta)
=-\sum_{t=1}^{T}\log P_\theta(s_t,q_{t+1}\mid Q,q_t).
\label{eq:transition-loss}
\end{equation}
Each observed transition supplies an independently batchable training instance. The data and computational pathway need not include the entire chain as one monolithic sequence. The result is a direct route from task structure to data parallelism, short local contexts, and reusable transition operators.
\begin{table}[t]
\centering
\caption{The pretraining paradigm changes when the primitive is a state transition.}
\label{tab:pretrain}
\small
\begin{tabular}{p{0.27\linewidth}p{0.30\linewidth}p{0.35\linewidth}}
\toprule
Dimension & General-purpose LLM & Markovian specialized model\\
\midrule
Attention & Global causal sequence & Local causal step + state interface\\
Training example & Long document / complete CoT & State--action--next-state triplet\\
Supervision & Joint next-token loss & Transition + state compression\\
Position & Global sequence coordinates & Reset step-relative coordinates\\
Gradient path & Long-sequence activations & Independent transitions / truncated recurrence\\
Parallelism & Sequence-aware TP/PP/CP & Strong data-parallel scaling\\
Evaluation & Perplexity and long retrieval & Transition, sufficiency, chain success\\
\bottomrule
\end{tabular}
\end{table}
\subsection{Dataset construction}
The critical operation is to cut each reasoning chain at \emph{semantically independent boundaries}, not at arbitrary token positions. From a trajectory $(q_1,s_{1:T},a)$, construct $(q_t,s_t,q_{t+1})$ samples; use a capable generator to produce reduced successor problems; then conduct a zero-history verification test: can a solver continue from $q_{t+1}$ without the earlier transcript? Reject or regenerate samples that fail. Counterfactual perturbations distinguish genuine state dependence from shortcuts that recover future steps from unrelated historical patterns.
Programs, automata, theorem-proving tactic states, symbolic algebra, chess-like game states, and tool-call traces admit large volumes of verifiable transitions. Different instances can be packed into longer hardware batches with block-diagonal attention masks that prevent cross-sample leakage. The data curriculum begins with explicit transitions and moves toward long autonomous state evolution.
\subsection{Multi-objective loss}
The native training objective joins transition prediction with constraints that make the state representation useful and minimal:
\begin{align}
\mc{L}
&=\mc{L}_{\mathrm{NTP}}+
\lambda_1\mc{L}_{\mathrm{trans}}+
\lambda_2\mc{L}_{\mathrm{bottleneck}}+
\lambda_3\mc{L}_{\mathrm{KD}}\nonumber\\
&\quad+\lambda_4\mc{L}_{\mathrm{dep}}+
\lambda_5\mc{L}_{\mathrm{drift}}+
\lambda_6\mc{L}_{\mathrm{boundary}}.
\label{eq:multi-objective}
\end{align}
Transition NLL teaches the local operator; an information bottleneck prevents the state from degenerating into a copied transcript; dependence losses enforce invariance to irrelevant earlier prefixes; full-context teacher distillation transfers useful conditional distributions into the state-only student; drift consistency compares free-running and directly initialized transitions; and boundary supervision recognizes step completion and chain termination.
The resulting model is \emph{trained to forget the irrelevant past}. Its memory economy is learned and architecturally enforced rather than applied as an afterthought.

\section{Dual-Loop Transformer Architecture}
\label{sec:architecture}
\subsection{Outer transition loop and inner depth loop}
\begin{figure}[t]
\centering
\includegraphics[width=\linewidth]{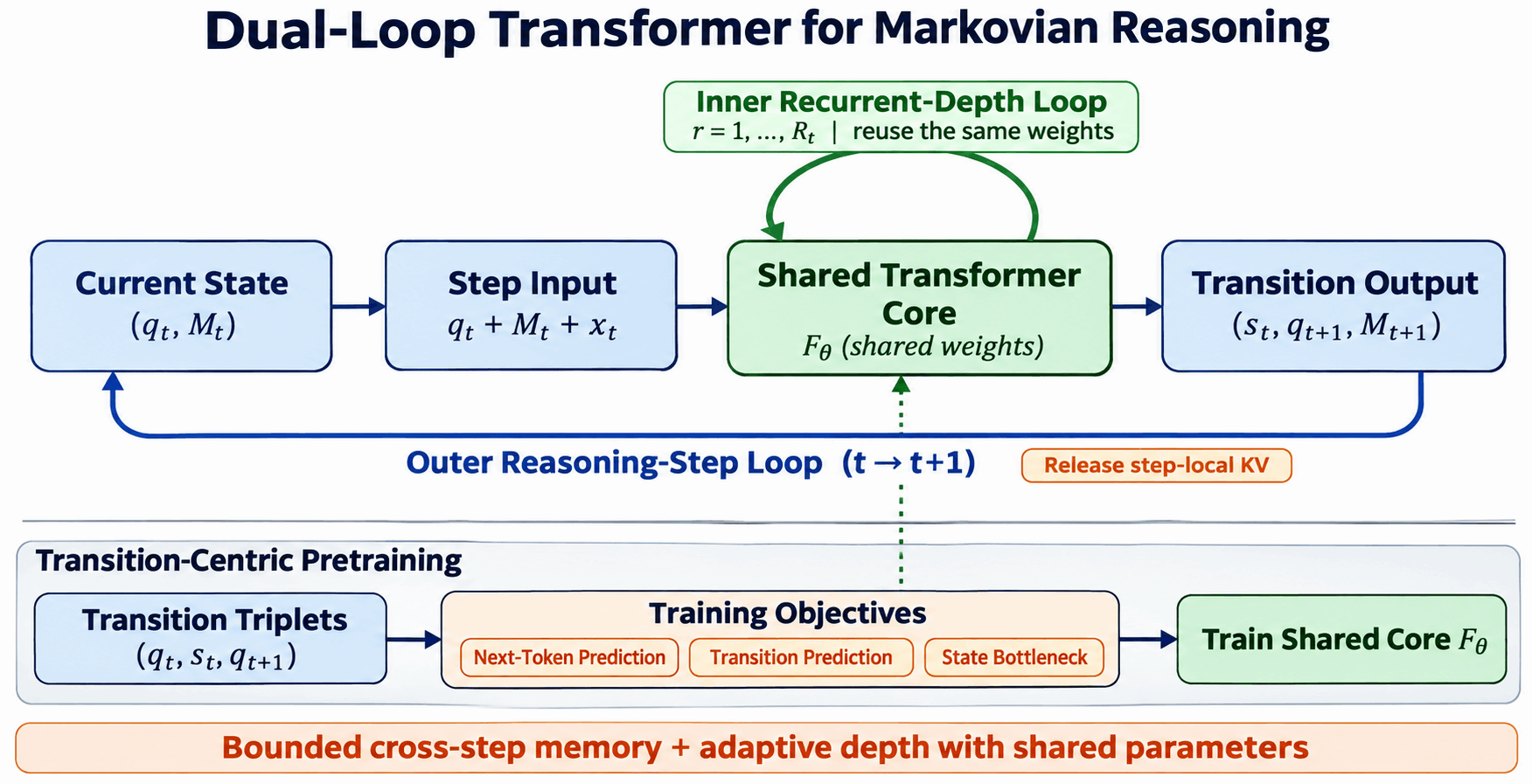}
\caption{Dual-loop Transformer architecture with transition-centric pretraining.
The outer loop advances the reduced state $(q_t,M_t)$ across reasoning steps
and releases step-local KV, while the inner loop repeatedly applies a shared
Transformer core for $R_t$ computation iterations. The transition training
objectives shape the same shared core to support state evolution and compact
cross-step memory.}
\label{fig:dual-loop}
\end{figure}

Two families of recurrent Transformers supply complementary building blocks. Recurrent Memory Transformers pass memory tokens between segments \citep{ref8}; Universal and looped Transformers reuse a computational stack at increasing effective depth \citep{ref9,ref11,ref24,ref25,ref32}. We integrate them into a dual-loop design:
\begin{itemize}
\item \textbf{Outer step loop ($t=1,\ldots,T$):} consumes current state $q_t$ and memory $M_t$, generates action $s_t$ and successor state $q_{t+1}$, and discards step-local KV.
\item \textbf{Inner depth loop ($r=1,\ldots,R_t$):} repeatedly applies a shared Transformer stack to the current step's hidden representation, allowing hard transitions to receive more computation without adding parameters.
\end{itemize}
The architecture uses a single shared transition core, as visualized in Figure~\ref{fig:dual-loop}. Depth is a reusable computational resource; the state is the sole cross-step semantic interface.

\subsection{Explicit memory tokens and recurrent matrix states}
An interface of $m$ memory tokens provides a learned sufficient statistic for the next step. Read-memory tokens precede each segment; write-memory tokens at its end are carried into the next segment. A small range such as $m=4$--$16$ is a useful design starting point, with capacity determined by the dimension of the task state. Prefix-token injection permits a nonintrusive wrapper around a pretrained Transformer, while read/write separation creates a direct differentiable path through the state interface.

As an alternative, a fixed-dimensional matrix recurrence can carry state in the style of delta-rule linear attention \citep{ref15,ref16,ref17}. With state matrix $S$, key $k$, value $v$, query $q$, decay $g$, and write factor $\beta$:
\begin{equation}
S\leftarrow S e^g,\quad
\hat v=k^\top S,\quad
\Delta=(v-\hat v)\beta,\quad
S\leftarrow S+k\Delta^\top,\quad
o=q^\top S.
\label{eq:delta}
\end{equation}
Memory tokens offer straightforward adaptation of existing weights; matrix states offer a compact native recurrent kernel.

\subsection{Adaptive computation and termination}
Simple state transitions may require one depth iteration, while difficult steps merit several. Options include fixed depth, a learned router assigning $R_t$ from a small set, and fixed-point halting based on
\begin{equation}
\rho_r=\frac{\lVert h^{(r+1)}-h^{(r)}\rVert_2}
{\lVert h^{(r)}\rVert_2+\delta}.
\label{eq:fixed-point}
\end{equation}
We favor fixed-point criteria with a hard $R_{\max}$ and a training distribution that samples recurrent depth. This makes computation an adjustable inference-time resource without changing parameter count. Step-level stopping uses a terminal predicate or learned $p_{\mathrm{stop}}$ and an explicit maximum number of outer steps.

\section{Training, Runtime, and Complexity}
\label{sec:training-runtime}
\subsection{Three-stage training curriculum}
The training sequence is deliberately progressive. \emph{Stage I: transition warm-up} learns the basic state-action map and step boundary format from short, supervised transitions. \emph{Stage II: Markovian invariance consolidation} introduces prefix perturbations, state pairing, bottleneck penalties, distillation, and drift losses until the model's behavior depends on the state rather than incidental history. \emph{Stage III: long-chain self-rollout} mixes teacher-forced and model-generated states, applies scheduled sampling, and trains the model to remain stable over many successive transitions.

Each observed transition can be trained independently, removing the need to backpropagate through the full chain length. Where differentiable memory crosses several steps, truncated backpropagation through time (BPTT) keeps a bounded gradient window while transmitting detached memory between segments. Within the depth loop, gradients still propagate across the recurrent applications; activation checkpointing and carefully scheduled recurrence depth control memory.

\subsection{Training stability as a dynamical-system problem}
Repeatedly applying one stack can amplify unstable hidden-state directions. A simplified recurrence
\begin{equation}
h^{(r+1)}=\alpha h^{(r)}+\beta e+F_\theta(h^{(r)})
\label{eq:stability}
\end{equation}
suggests the importance of Jacobian spectral control, residual scaling, pre-normalization, gradient clipping, step-wise input reinjection, smaller learning rates for recurrent parameters, and curricula that gradually increase loop depth. These controls prevent hidden-state collapse, spurious early fixed points, and the failure mode in which additional iterations cease improving the answer.

\subsection{Inference engine}
The serving stack contains a step scheduler, input constructor, recurrent core, sampler, state manager, KV block manager, and Markovian validity monitor. Three practical cache policies are available: clear and rebuild step-local KV at each boundary; store a fixed state register that survives across steps; or project an explicit dense state into a few virtual memory tokens. A hybrid implementation can compare compact and full-context paths on sampled steps and invoke a fallback when their distributions diverge.

For branch search, memory tensors can be replicated and advanced in batches rather than copying unbounded historical KV. The runtime can schedule multiple branches, score successor states, retain the best candidates, and checkpoint the small state at selected steps.

\subsection{Asymptotic cache and compute}
Let $N$ denote Transformer layers, $H$ KV heads, $D$ head dimension, $b$ bytes per scalar, $P$ static prompt tokens, $s$ sink tokens, $m$ memory tokens, $L_i$ the length of step $i$, and $L_{\max}$ a step-length budget. Full-history and state-structured cache allocations are respectively
\begin{align}
M_{\mathrm{full}}
&=2NHDb\left(P+\sum_{i=1}^{k}L_i\right)
=O(k\bar L),\label{eq:fullmem}\\
M_{\mathrm{Markov}}
&=2NHDb\left(P+s+m+L_{\max}\right)
=O(1)\quad\text{in }k.\label{eq:markovmem}
\end{align}
Thus the number of \emph{reasoning steps} no longer determines memory allocation. With shared weights, effective computational depth increases as $R_tL_c$ while parameter count remains that of the $L_c$-layer shared core. The design converts long-horizon memory growth into a bounded state and an adjustable computational budget.

If an approximate state transition has independent step error rate $\epsilon$, the probability that all $k$ steps are correct follows the indicative relationship $(1-\epsilon)^k$. State validation, occasional recomputation anchors, and branch checkpoints directly address this long-horizon error propagation.

\section{Evaluation Protocol and Initial Feasibility Evidence}
\label{sec:evaluation}
\subsection{Task-specific measurements}
Long-context retrieval benchmarks test an ability that the proposed architecture intentionally replaces with sufficient-state transfer. The central evaluation quantities are:
\begin{itemize}
\item \textbf{Transition fidelity:} one-step accuracy and KL divergence from a full-context teacher.
\item \textbf{State sufficiency:} success when the next stage receives only the reduced state and fixed task condition.
\item \textbf{Chain success and extrapolation:} final correctness as $k$ increases well beyond training horizons.
\item \textbf{Counterfactual robustness:} correct response to perturbations of the current state, independent of irrelevant prefixes.
\item \textbf{Cache--horizon decoupling:} measured KV footprint, peak memory, time per output token, and their dependence on $k$.
\item \textbf{Compute efficiency:} accuracy as a function of the number of inner-loop iterations.
\end{itemize}
Baselines include full-context LLMs, bounded-history cache eviction, sliding-window attention, SSM or linear-attention cores, and fixed-depth specialized Transformers. Synthetic transitions, mathematical reasoning, program execution, theorem proving, and process validation should be represented jointly.

\subsection{Controlled state-only inference probe}
To quantify the immediate benefit of a sufficient-state interface, we conducted a controlled inference probe on three pretrained backbones. For each model, deterministic addition and modulo-10 transition tasks were evaluated at history lengths $k\in\{1,8,32,128\}$. A full-history prompt and a current-state-only prompt received identical current states and next operations. We measured \emph{actual KV tensor sizes} in a PyTorch forward pass and recorded greedy generation outcomes. This prompt-level experiment directly demonstrates the memory advantage of state transmission before architecture-specific training.
\begin{table}[t]
\centering
\caption{Controlled 128-step probe, 12 paired cases per task and model. Cache entries are mean measured KV tensor sizes (MiB); accuracy is number correct out of 12. Full retains all earlier transition text and State carries only the current value.}
\label{tab:pilot}
\scriptsize
\begin{tabular}{llrrrr}
\toprule
Backbone & Task & Full KV & State KV & Full acc. & State acc.\\
\midrule
Qwen3-8B & Addition & 333.54 & 8.16 & 10/12 & 8/12\\
Qwen3-8B & Modulo 10 & 336.23 & 8.44 & 10/12 & 7/12\\
Qwen2.5-Coder-14B-Instruct & Addition & 447.91 & 14.06 & 12/12 & 12/12\\
Qwen2.5-Coder-14B-Instruct & Modulo 10 & 451.50 & 14.44 & 12/12 & 12/12\\
Mistral-7B-Instruct-v0.3 & Addition & 296.35 & 7.00 & 11/12 & 12/12\\
Mistral-7B-Instruct-v0.3 & Modulo 10 & 298.88 & 7.38 & 5/12 & 11/12\\
\bottomrule
\end{tabular}
\end{table}

Across all three backbones, state-only KV size stays essentially constant as the history grows, while the full-history baseline grows with the transcript. On Qwen2.5-Coder-14B-Instruct, the 128-step state interface reduced measured cache size by approximately $96.9\%$ on addition while maintaining perfect correctness in all 12 cases. The cross-backbone results establish a compelling direction: turn an already powerful state-compression interface into a native model capability through transition-centric training and Markovian invariance. Appendix~\ref{app:probe} provides the complete per-model results and timing details.

\section{Discussion and Vision}
\label{sec:discussion}
The proposed specialization makes a decisive architectural choice: long-horizon reasoning should be built around \emph{state evolution}, not the mechanical accumulation of its entire textual trace. A well-designed state carries every constraint needed by the next transition, while the serving engine avoids allocating memory to redundant completed steps. In more complex workflows, the same principle naturally extends to richer state carriers and hybrid state interfaces, preserving the model's state-first organization.

State-space models compress through recurrence; our formulation adds a task-level criterion for what the recurrence must preserve. The two ideas are complementary. A promising endpoint is a hybrid in which an SSM serves the outer Markovian loop and a Transformer serves the inner depth loop. An equally compelling system-level arrangement delegates global planning and open-domain interaction to a general-purpose language model, while a specialized recurrent kernel executes long sequences of structured transitions and returns the terminal state.

Neuro-symbolic verification strengthens this state-centered vision. LogicSAGE couples an 8B neural reasoner with Socratic-guided symbolic correction, demonstrating the power of specialized reasoning and explicit feedback \citep{ref38}. The probabilistic soft-logic verifier ProbVeri extends this principle to uncertainty-bearing reasoning steps through contrastive energy-based inference \citep{ref39}. Both directions complement our execution architecture: a compact successor state can be inspected by symbolic or probabilistic verification modules, and their feedback can shape the next transition without requiring an ever-growing textual history.

This division of labor extends beyond mathematical CoT. It invites compact reasoning machines for symbolic computation, tool-mediated agents, theorem provers, industrial workflows, automated verification, and any domain where a well-defined state and a repeatable transition operator exist. The next research frontier is to learn \emph{adaptive sufficient-state capacity}: not one fixed memory size for every problem, but a state interface that grows only when the true dimensionality of the task demands it.

\section{Conclusion}
\label{sec:conclusion}
We present a unified design for Markovian-reasoning specialized language models, connecting formal state sufficiency, structural KV-cache pruning, transition-centric pretraining, and a dual-loop recurrent Transformer. The core consequence of a fixed sufficient-state interface is decisive: cache size becomes independent of the number of completed reasoning steps, while shared recurrent weights allow computation depth to be assigned where each transition needs it. The training objective makes this behavior native; the inference engine turns it into practical step-wise state management; and the evaluation protocol measures the qualities that matter for this task family.

\textbf{The guiding vision is a shift from remembering as much as possible to transmitting only what is necessary.} Specialized reasoning can be long-horizon without being long-context. Sufficient-state transmission, step-parallel training, and parameter-shared recurrent computation together define a powerful alternative to the ever-growing reasoning transcript. We envision a family of specialized large language models whose reasoning horizon grows through repeated state transitions while memory remains bounded, computation remains adaptable, and every step is organized around an explicit state of the problem.

\appendix
\section{Detailed Cache Design and Engineering Realization}
\label{app:cache}

\subsection{Why a standard decoder retains historical KV}
In a conventional causal Transformer, the query of token $t$ may attend to the key/value vectors of every earlier token, so the cache grows unboundedly as decoding proceeds. In a Markovian task, the two roles played by attention can be separated. The first is \emph{semantic aggregation}: recovering useful information from earlier steps. The second is \emph{nonsemantic attention support}: retaining stable positions that accumulate attention mass, including attention sinks. Once all task-relevant history is captured in $z_i$, the semantic role no longer requires raw earlier-step tokens, while a fixed number of sink tokens can supply the second role.

The static condition $Q$ must be distinguished from historical steps. It may include the system directive, the original problem, and persistent rules. It remains present throughout the chain, can be prefix-cached across requests when identical, and is never removed merely because an earlier reasoning segment has been completed. The cache contract is
\[
\ctx_i=\mathrm{KV}(Q)\oplus\mathrm{KV}(\mathrm{sink})\oplus
\mathrm{KV}(z_i)\oplus\mathrm{KV}(\mathrm{current\ step}).
\]
The cost of the fixed prompt is not a cost that grows with the number of reasoning transitions.

\subsection{Training-free, training-aligned, and recurrent alternatives}
\begin{table}[h]
\centering
\caption{Implementation routes for Markovian cache pruning.}
\small
\begin{tabular}{p{0.18\linewidth}p{0.45\linewidth}p{0.29\linewidth}}
\toprule
Route & Mechanism & Precision and engineering cost\\
\midrule
A: Zero-training approximation & Step-wise rolling of sinks, static prompt, and current state under unchanged weights & Approximate; divergence probes and fallback\\
B: Training-aligned masking & Block-level causal mask, step-end state token, local position convention, SFT & Exact computation graph; training pipeline changes\\
C: Architecture replacement & Fixed-size SSM or linear-attention recurrence & Native bounded state; exact historical retrieval is not its goal\\
\bottomrule
\end{tabular}
\end{table}
Route A is a fast way to assess whether a pretrained model behaves as though the task were approximately Markovian. Route B is the direct means of making the retention rule part of the model definition. Route C is the native end-state architecture. A practical research sequence uses A to audit the assumption, B to establish aligned transitions, and C to distill the computation into a recurrent kernel.

Conditional independence in the data-generating process is not automatically the attention graph of a conventional pretrained Transformer. Dropping earlier KV only at inference can change attention normalization and RoPE geometry. The training-aligned route establishes a block-causal mask and position convention common to training and decoding.

\subsection{The three cache regions}
The static region stores the system/problem prefix and a small attention-sink set. The dynamic state region stores $m$ memory tokens or a fixed matrix recurrence. The within-step region stores causal KV for the tokens currently being generated. A state-slot rotation occurs at every semantic step boundary:
\begin{enumerate}
\item Begin step $i$ with the static region, $z_{i-1}$, and an empty step-local buffer.
\item Generate the current action and reduced successor problem with ordinary causal attention \emph{within that step}.
\item Extract the new sufficient state $z_i$ from designated write-memory positions.
\item Release the step-local KV and atomically replace the state region by $z_i$.
\item Restart the local positional coordinate system and proceed to step $i+1$.
\end{enumerate}
For a state register, each layer/head can reserve a fixed-capacity slot for $m$ state vectors, while the current step uses a bounded ring or block buffer. This separates state overwrite from ordinary within-step causal decoding. The design avoids per-step growth in the block table.

\begin{quote}
\small\noindent\textbf{Algorithm: Step-wise bounded-state decoding}\label{alg:decode}\\
Initialize $(Q_{\rm KV},K_{\rm sink},z_0)\gets\mathrm{PrefillAndReduce}(Q)$.\\
\textbf{For} each step $i=1,\ldots,k$:\\
\hspace*{1em}Clear local buffer $B_{\rm step}$; set local position to zero.\\
\hspace*{1em}\textbf{While} the step boundary has not been reached:\\
\hspace*{2em}Decode with $(Q_{\rm KV},K_{\rm sink},z_{i-1},B_{\rm step})$.\\
\hspace*{2em}Advance local position and update step-local buffer.\\
\hspace*{1em}Extract $z_i$ using the read/write memory interface.\\
\hspace*{1em}Release $B_{\rm step}$ and replace the state register.\\
\hspace*{1em}\textbf{Break} if $\mathrm{Term}(z_i)$.
\end{quote}

\subsection{Position, attention sinks, and serving blocks}
The logical positions of tokens must not be confused with their physical cache slots. Under RoPE, an overwritten slot cannot simply inherit the slot index as its token position; such a change alters relative-position behavior. We recommend local position indices beginning at zero for each step, fixed virtual positions for the state interface, and an explicit boundary-reset rule. When using global absolute positions in an existing checkpoint, the original logical positions must instead be preserved.

Attention sinks, commonly a small fixed set of four to eight tokens, can remain pinned in a ring buffer. With PagedAttention \citep{ref6}, the state register can be mapped to reserved physical blocks and updated at fixed offsets. The allocator then maintains a stable block budget rather than continuously appending historical step blocks. Step-local blocks are released at the semantic boundary. Block-sparse or variable-length IO-aware kernels \citep{ref19} implement the attention mask while keeping computation efficient.

\subsection{Checking dependency range before pruning}
Four complementary probes establish the state interface:
\begin{enumerate}
\item \textbf{Distant-prefix intervention.} Inject information at an earlier step $j<i-1$ and measure whether the next-step output distribution changes when the final state is fixed. KL or Jensen--Shannon distance provides a direct numerical signal.
\item \textbf{Layer/head attention inspection.} Estimate the fraction of attention mass outside a local window of width $d$. A per-head or per-layer analysis guides hybrid local/global configurations.
\item \textbf{Counterfactual state intervention.} Perturb $s_i$ or $z_i$ and check whether the successor changes according to the transition rule. The model must use the state rather than recite a memorized trajectory.
\item \textbf{Mixed-depth fallbacks.} Keep a few global bypass heads or a bounded local window when a task has weak residual long-distance dependencies. This creates a smooth transition between exact and approximate state sufficiency.
\end{enumerate}

\subsection{Checkpoints, rollback, and branching}
In-place overwrite removes raw preceding states. A search procedure that explores branches therefore maintains explicit checkpoints. Every $N$ steps, serialize the state register and logical position metadata to CPU memory or NVMe. Restoring a branch loads the nearest checkpoint and replays the required transitions. State hashes and token/position identifiers detect branch misalignment. The footprint of one checkpoint depends on the fixed memory interface, not the full generated transcript.

\subsection{Strategy selection}
When a task is proven or strongly verified to be Markovian, the structural state-slot route gives the strongest memory bound. For approximately Markovian problems with uncertain distant dependencies, a small window combined with recurrent attention is a natural fallback. For open-domain tasks dominated by arbitrary historical retrieval, attention-score eviction and transfer compression remain valuable tools. The objective is to select a model architecture matched to the dependency structure of the workload.

\section{Complete Transition-Centric Learning Recipe}
\label{app:pretraining}

\subsection{How specialized pretraining differs from general language modeling}
General-purpose LLM pretraining rewards prediction under arbitrary long-range token dependencies. In the specialized setting, each example teaches a conditional transition from $q_t$ to $(s_t,q_{t+1})$. A short local context, often on the order of hundreds to a few thousand tokens depending on task-state complexity, replaces a long chain as the elementary sample. The objective need not require extrapolation of global positional embeddings across the entire reasoning horizon.

The transition-centric model consequently places less emphasis on long-context extension techniques and more emphasis on a dedicated state compression head, state sufficiency, and robust step boundaries. A recurrent or sliding-window backbone is appropriate when its training and inference windows agree. Mamba-style selective state models provide a natural fixed-size carrier, but being a state-space model alone does not force its hidden state to forget irrelevant history; the transition and bottleneck objectives are what impose that specific inductive bias.

\subsection{Constructing state--action--next-state triplets}
The data pipeline follows seven operations:
\begin{enumerate}
\item \textbf{Split the reasoning trajectory.} Map $(q_1,s_{1:T},a)$ into $T$ independent triplets $(q_t,s_t,q_{t+1})$.
\item \textbf{Generate reduced problems.} A strong teacher produces $q_{t+1}$ as a new problem whose continuation can be solved without the earlier derivation.
\item \textbf{Verify independence.} A separate solver receives only $q_{t+1}$ and attempts the remaining task. Reject or regenerate failures. This is a substantive sufficiency check, not a string-length check.
\item \textbf{Remove long-range shortcut signals.} For task-focused pretraining, favor self-contained segments over arbitrary long-document concatenations that encourage the model to exploit unavailable history.
\item \textbf{Pack samples with block-diagonal masks.} Independent short samples can be packed into efficient hardware sequences, but cross-example attention must remain disabled.
\item \textbf{Augment counterfactually.} Perturb states and generate their correct successors, explicitly breaking spurious historical correlations.
\item \textbf{Cut at natural boundaries only.} Token-length truncation is not semantic reduction. Boundaries should coincide with a new mathematical subproblem, tool-call round, tactic state, or verified state-machine transition.
\end{enumerate}
Program execution traces, finite automata, theorem-proving tactic states, game records, symbolic algebra, and tool usage provide procedurally generated transitions with exact labels. Small high-quality sets can be used to adapt pretrained models before scaling the specialized corpus.

\subsection{Detailed objective components}
\begin{description}
\item[Transition NLL.] Minimize negative log-likelihood of $(s_t,q_{t+1})$ conditioned on $(Q,q_t)$. Within the step, the causal mask prevents leakage from the target reduced state into the generated action.
\item[State bottleneck.] Constrain the length or entropy of $q_{t+1}$ and penalize copies of irrelevant historical text. A contrastive term can distinguish minimally sufficient states from longer, redundant alternatives. An auxiliary adversary can attempt to reconstruct older history from the state, with its success penalized subject to preservation of next-step predictive information.
\item[Independence consistency.] Train a discriminator or verifier to judge whether a proposed successor state is sufficient for solving the rest of the task. Correct and incorrect successor pairs can also support preference optimization.
\item[Full-context teacher distillation.] Match the teacher's next-action logits using a student that receives only $z_i$. This transfers predictive information through the bottleneck.
\item[Drift consistency.] Compare $S_1\rightarrow S_2\rightarrow S_3$ with a direct one-step transition initialized from the same $S_2$. A KL term encourages the same operator regardless of the path used to obtain the state.
\item[Loop consistency.] Compare a fixed state processed for $R$ and $R+1$ inner iterations; prevent extra computation from becoming uncontrolled stochastic resampling.
\item[Boundary recognition.] Train step-end markers, transition markers, and chain-end markers so the outer loop operates on the correct semantic units.
\end{description}
The multi-objective construction in Eq.~\eqref{eq:multi-objective} coordinates these terms. The goal is neither an empty state nor a copied transcript: it is a representation carrying precisely the information necessary for the successor transition.

\subsection{Three stages of pretraining and adaptation}
The first stage retains a reduced general-language foundation, sufficient to parse natural-language descriptions of states. It uses shorter sequence lengths and does not invest in long-context extension merely to support an eventual unbounded reasoning transcript. The second stage is state-transition pretraining: synthetic Markovian tasks train the model to compress intermediate reasoning into a self-contained successor problem and predict the next transition. The third stage is supervised task adaptation using MCoT-like triplets, followed by self-distillation and longer generated trajectories.

Because independent state-transition pairs can be shuffled freely, data parallelism is especially effective. Model and pipeline parallelism remain available when parameter size demands them, but context parallelism no longer carries the burden of processing one extremely long chain. The desired engineering effect is higher sample throughput and a shorter gradient path for the same number of supervised transitions.

\section{Detailed Dual-Loop Realization}
\label{app:dual-loop}

\subsection{Comparing recurrent design lines}
\begin{table}[h]
\centering
\caption{Two complementary recurrent Transformer traditions.}
\small
\begin{tabular}{p{0.23\linewidth}p{0.34\linewidth}p{0.35\linewidth}}
\toprule
Dimension & Token/segment memory (RMT, RATE) & Recurrent depth (MoR, Ouro)\\
\midrule
Recurrence meaning & Carry state between segments & Reapply one stack to hidden states\\
State carrier & Learnable memory tokens & Updated hidden representations\\
Cross-step KV & Memory-token KV & Cleared at step boundaries\\
Single-step compute & One stack application & Multiple shared-weight applications\\
Characteristic risk & Memory capacity bottleneck & Cache/position errors or loop collapse\\
\bottomrule
\end{tabular}
\end{table}
Our recommended realization uses recurrent depth as the computational backbone and memory tokens as the explicit state interface. The outer loop advances the Markovian trajectory; the inner loop supplies the depth required to compute a difficult transition. The same shared transition operator is reused because each step represents another application of the same conditional rule.

\subsection{Algorithmic skeleton}
\begin{quote}
\small\noindent\textbf{Algorithm: Dual-loop Transformer with adaptive depth}\label{alg:dual}\\
Initialize state $q_1$, memory $M_1$, and shared stack $F_\theta$.\\
\textbf{For} $t=1,\ldots,T_{\max}$:\\
\hspace*{1em}Set $h^{(0)}\gets\mathrm{Embed}([\mathrm{BOS}],q_t,M_t)$.\\
\hspace*{1em}\textbf{For} $r=1,\ldots,R_{\max}$:\\
\hspace*{2em}Update $h^{(r)}\gets F_\theta(h^{(r-1)},e_r)$.\\
\hspace*{2em}\textbf{Break} on fixed-point convergence.\\
\hspace*{1em}Decode $(s_t,q_{t+1},M_{t+1})$ from the final hidden state.\\
\hspace*{1em}Reset step-local KV; \textbf{break} if $\mathrm{Term}(q_{t+1})$.
\end{quote}
The inner loop can use learned loop embeddings or another explicit recurrence-depth coordinate. Its hidden-state updates and shared weights determine effective depth; the step loop determines the number of state transitions. Consequently, increasing reasoning horizon does not increase the number of learned parameters or retained historical KV vectors.

\subsection{Memory token quantity, read/write routing, and injection}
The number $m$ of memory tokens is an explicit information bottleneck. A practical starting range is $4$--$8$ for compact numerical states and $8$--$16$ for structured automata, increasing to $16$--$32$ when sufficiency probes reveal lost constraints. Read tokens appear at segment input and write tokens at segment output. The preceding write slots become the subsequent read slots, retaining a differentiable path through a small interface.

Prefix-token injection is the simplest wrapper around pretrained Transformers. At each inner iteration, reinject the original step embedding as a stabilizing signal. Two particularly important recurrent failure modes are \emph{hidden-state collapse}, in which token representations become nearly identical, and \emph{loop collapse}, in which the model learns to ignore recurrence and gives the same output after one or many iterations. Sandwich normalization, careful output-projection initialization, residual scaling, lower learning rates, and intermediate supervision counter these failures.

\subsection{Matrix-state alternative and numerical stability}
The delta-rule recurrence in Eq.~\eqref{eq:delta} corrects an existing state prediction rather than blindly accumulating repeated writes. To maintain numerical stability over long chains, normalize keys, control the update magnitude (for example, keep $\beta\lVert k\rVert^2$ within a stable range), accumulate sensitive state statistics in FP32, and periodically rescale blocks. Memory-token and matrix-state variants offer complementary implementation paths: the former is accessible through pretrained Transformer wrappers, while the latter gives a compact inference-oriented recurrent kernel.

\subsection{Adaptive loop depth}
Three depth-control mechanisms are natural:
\begin{enumerate}
\item \textbf{Fixed $R$.} Choose a constant such as $R=2$ or $R=4$ for predictable latency and memory.
\item \textbf{Fixed-point / adaptive computation time.} Stop when normalized hidden-state change falls below a threshold, up to $R_{\max}$ \citep{ref10}.
\item \textbf{Learned routing.} A step-level router selects a depth from a small set (for example $\{1,2,4,8\}$), and token-level routing may stop easy tokens earlier \citep{ref25}.
\end{enumerate}
Randomized depth sampling during training prevents one fixed iteration count from becoming hardwired into the parameters. Heavy-tailed sampling allocates most training updates to short recurrences while occasionally exposing the model to deeper computation. Training and inference must share the relevant recurrence-depth support; inference should not silently exceed the depth range that the training dynamics have learned.

\section{Progressive Training and Recurrent Stability}
\label{app:training}
\subsection{Three training stages in detail}
\paragraph{Stage I: transition-operator warm-up.} This stage establishes reliable local mapping before imposing the full Markovian bias. Use abundant single-step transition pairs and short trajectory fragments. Train with transition cross-entropy and boundary-recognition losses; keep invariance and bottleneck penalties at zero or a very small weight. Monitor single-step perplexity, exact transition success, and correct recognition of the step-end marker. The objective is to make the recurrent unit competent at the fundamental state-to-successor computation.

\paragraph{Stage II: consolidation of Markovian invariance.} This is the central transformation from a general sequence model into a state-transition model. Introduce paired transitions whose current state is identical but whose distant history differs, randomized irrelevant prefixes, and positional perturbations. Increase history-invariance, state-alignment, information-bottleneck, and drift-consistency penalties progressively. Track the \emph{Markovian invariance rate}: the fraction of matched-state pairs whose successor answers or distributions agree within tolerance. Also measure state fidelity, past-history predictability from encoded states, correct forgetting after state updates, and boundary accuracy. A practical target is invariance and fidelity above a threshold such as $95\%$, provided one-step transition success remains intact. A model that merely loops is not yet a Markovian reasoner; the intended inductive bias is that its future depends on explicit sufficient state rather than incidental prior text.

\paragraph{Stage III: long-horizon autonomous rollout.} Mix ground-truth and model-generated states through scheduled sampling. Preserve transition, bottleneck, invariance, distillation, and boundary objectives while emphasizing rollout consistency. Pure teacher forcing is insufficient because the model never learns to recover from its own small transition errors. Long-horizon training must expose the operator to the states it generates in deployment.

\subsection{Truncated BPTT and gradient accounting}
The factorization in Eq.~\eqref{eq:factorization} permits independently observed transitions to be shuffled and trained in batches. Full $T$-step backpropagation is unnecessary for supervised one-step prediction. For differentiable state transfer over multiple outer steps, divide a trajectory into length-$B$ windows and detach memory at each boundary; the memory transports information, while the backward graph remains bounded by $B$. Within each step, retain gradients through the selected $R$ shared-weight depth applications, using activation checkpointing to control retained activations. During reinforcement-learning post-training, reconstruct memory and hidden states with the current policy parameters rather than reusing stale rollout-time tensors, maintaining consistency of the differentiated action distribution.

\subsection{The repeated core as a dynamical system}
A simplified recurrence,
\begin{equation}
h_{r+1}=\alpha h_r+\beta e+F_\theta(h_r),
\end{equation}
illustrates how repeated application can amplify unstable directions. A locally linearized update with spectral radius greater than one can multiply perturbations exponentially with depth. Pre-normalization, residual scaling, constraints on injection projections, gradient clipping, smaller recurrent-core learning rates, and a depth curriculum such as $2\rightarrow4\rightarrow8$ stabilize the iteration. Intermediate supervision counters hidden-state collapse and makes additional iterations meaningful. Test explicitly for insufficient memory capacity, false convergence or loop collapse, accidental accumulation of step KV, inconsistent positional indices, and inference depth outside the training support.

\section{Inference-Engine Realization}
\label{app:runtime}
\subsection{Prefill, recurrent execution, and adaptive termination}
The runtime contains a loop scheduler, input constructor, shared recurrent core, sampler, state manager, KV-block manager, and Markovian validity checker. Following prompt prefill, only the fixed static condition, selected attention sinks, and current state interface survive. Static $Q$ tokens may be reused by prefix caching. Each outer step builds input from static condition, state memory, and a bounded local workspace; executes $R_t$ recurrent-depth updates; samples the local segment and successor; and replaces the state slot at the boundary. Termination follows the task predicate, learned stopping probability, fixed-point condition, or configured $T_{\max}$. Since parameter count is independent of recurrent depth, a serving scheduler may shift computation toward difficult steps without loading different model weights.

\subsection{Three cache policies}
\paragraph{Clear-and-rebuild.} At each step, reconstruct a fresh local KV allocation and release it after extracting the successor state. This provides a simple semantically transparent implementation but repeats encoding.

\paragraph{Fixed recurrent register.} Preserve a fixed-size register containing static prefix KV and the latest state KV; reset or overwrite transient blocks. The bounded register is a practical deployment default compatible with fixed-size GPU memory pools.

\paragraph{Virtual-state cache.} Carry a dense latent state and project it into a handful of virtual tokens before each transition. Its storage approaches a fixed constant and its effectiveness depends on trained state fidelity. A hybrid policy compares state-only and recent-segment paths on selected steps, falling back to longer context when output distributions diverge.

\subsection{Parallel branching with explicit states}
The current state is a small tensor, so a batch of $B$ candidate trajectories requires a memory bank of shape $[B,m,d]$, not $B$ copies of the entire old transcript. Advance candidates together, score successors, select the top $K$ branches, and release all local KV. Checkpoints at chosen boundaries save a fixed state and positional metadata, enabling replay after a failed continuation.
\begin{quote}
\small\noindent\textbf{Algorithm: Batched dual-loop state branching}\\
\textbf{Input:} branch states $Q_0$, memory bank $M_0$, depth limit $R_{\max}$.\\
\textbf{For} $t=0,\ldots,T_{\max}-1$:\\
\hspace*{1em}Set $H\gets\mathrm{Embed}(Q_t,M_t)$.\\
\hspace*{1em}\textbf{For} $r=1,\ldots,R_{\max}$:\\
\hspace*{2em}Set $H_{\mathrm{old}}\gets H$ and $H\gets F_\theta(H,\mathrm{LoopEmb}(r))$.\\
\hspace*{2em}\textbf{Break} if $\mathrm{Converged}(H,H_{\mathrm{old}})$.\\
\hspace*{1em}Decode $(Q',M',w)$ and select top-$K$ successor states.\\
\hspace*{1em}Release step-local KV and update $(Q_{t+1},M_{t+1})$.\\
\hspace*{1em}\textbf{Break} if every branch is terminated.
\end{quote}

\subsection{Validation and fallback}
Validation hooks check step boundaries, irrelevant-prefix invariance, compressed-state fidelity against a longer-context teacher, long-horizon drift, valid termination, and GPU memory allocation. Checks can be periodic or sampled. Failure can trigger an explicit fallback from compressed state to recent-segment input, from recurrent register to full re-encoding, or to an earlier checkpoint. The key engineering requirement is that failed state sufficiency be recognized and corrected, not hidden in the next transition.

\section{Complexity and Error Propagation}
\label{app:complexity}
With $N$ Transformer layers, $H$ KV heads, head dimension $D$, and $b$ bytes per scalar, each cached token consumes $2NHDb$ bytes. For static prompt length $P$, individual step lengths $L_i$, $s$ sink tokens, $m$ state tokens, and bounded current step $L_{\max}$:
\begin{align}
M_{\mathrm{full}}(k)&=2NHDb\left(P+\sum_{i=1}^{k}L_i\right),\\
M_{\mathrm{structured}}(k)&\leq 2NHDb(P+s+m+L_{\max}).
\end{align}
For a sufficient fixed-dimensional state and bounded local step, structured memory is independent of completed step count $k$, whereas full-history KV grows with cumulative token length. Quantization and low-rank latent KV can reduce constant factors further. Shared recurrent weights keep parameter storage at that of an $L_c$-layer core while offering effective depth $R L_c$; extra iterations still consume compute. The essential benefit is asymptotic decoupling of history length and cache budget, alongside a tunable compute budget.

If each independent transition succeeds with probability $1-\epsilon$, exact completion over $k$ transitions has indicative probability $(1-\epsilon)^k$. More generally, a state-distance recurrence can be bounded as $e_{t+1}\leq\alpha e_t+\delta_t$, where $\alpha$ is transition sensitivity and $\delta_t$ local approximation error. Contractive transitions control drift, while expansive transitions require stronger fidelity, periodic recomputation anchors, or rollback. This motivates reporting complete-chain success and error accumulation in addition to one-step accuracy.

\section{Experimental Program and Measurement Protocol}
\label{app:experiments}
\subsection{Evaluation criteria}
Evaluation should measure (1) transition fidelity and KL against a full-context teacher; (2) sufficiency of the reduced state; (3) final chain-level success; (4) length extrapolation when trained on short chains and tested at $k=100$ or greater; (5) correct reaction to counterfactual state changes with invariance to irrelevant prefixes; (6) self-correction after injected intermediate-state errors; (7) KV payload, peak GPU memory, and time per output token versus horizon; and (8) accuracy as a function of actual inner-loop depth. Conventional long-document retrieval benchmarks assess a different design objective.

\subsection{Baselines, datasets, and ablations}
Include four baseline families: full-KV general LLMs; cache-eviction and cache-compression methods such as H2O, SnapKV, and StreamingLLM; sliding-window attention, Mamba/SSM, and linear-attention state kernels; and specialized fixed-depth Transformers without the inner recurrent loop. Match backbone and evaluation conditions wherever a controlled architectural comparison is possible.

Build verifiable datasets from synthetic arithmetic and symbolic rewriting, deterministic automata, program traces, and rule-based state transitions. Add reduced-state triplets derived from GSM8K/MATH-style problems, theorem-proving tactic states, industrial process checks, structured compliance tasks, and domain-specific stepwise deduction. Controlled finite-state tasks are especially valuable because the true state dimension and transition kernel are known. Jointly ablate Markovian penalty weights, counterfactual-prefix perturbations, stopping policy, depth $R$, state size $d_s$, hard or soft masks, sink and memory-token counts, local segment length, and truncated-BPTT window. A joint state-capacity/depth sweep reveals the essential compression--computation tradeoff.

\begin{table}[h]
\centering
\caption{Recommended implementation roadmap, preserving the original manuscript's five-stage validation logic.}
\small
\begin{tabular}{p{0.08\linewidth}p{0.38\linewidth}p{0.43\linewidth}}
\toprule
Stage & Implementation & Acceptance signal\\
\midrule
0 & Audit remote historical dependence & Distant history is dispensable\\
1 & Training-free rolling KV with KL probes & Bounded distribution drift\\
2 & Block-causal mask, state tokens, multi-objective tuning & Stable transition accuracy\\
3 & Distillation into a native recurrent-state core & Long-horizon extrapolation\\
4 & Counterfactual, memory, TPOT, drift validation & Accuracy and resource targets met\\
\bottomrule
\end{tabular}
\end{table}
The minimal starting sequence is two- and three-step synthetic transitions, $m=8$ memory tokens and randomized $R\in[1,8]$, single-transition cross-entropy, extrapolation toward 50 steps, counterfactual probes, and finally replacement of the memory-token interface by a compact recurrent kernel.

\section{Controlled Three-Backbone Prompt Probe}
\label{app:probe}
This supplementary measurement isolates the \emph{state-only inference interface}. It is not the result of training the proposed dual-loop model; it measures an immediately testable consequence of replacing a long transition transcript with its current sufficient scalar state.
\subsection{Setup}
The pretrained backbones were Qwen3-8B, Qwen2.5-Coder-14B-Instruct, and Mistral-7B-Instruct-v0.3. Two deterministic scalar-state tasks---addition and modulo-10 arithmetic---used 12 paired cases at history lengths $k\in\{1,8,32,128\}$. Full-history prompts included the previous transition text; paired state-only prompts supplied the same current value and next operation without old text. Greedy generation produced answers. An actual forward pass measured key/value tensor element counts converted to MiB, and median generation latency was recorded in milliseconds.
\begin{table}[h]
\centering
\caption{Measured 128-step KV and generation-time comparison; 12 paired cases per row. Full and State KV are mean MiB; times are median ms. Accuracy appears in Table~\ref{tab:pilot}.}
\label{tab:pilot-times}
\scriptsize
\begin{tabular}{llrrrrr}
\toprule
Backbone & Task & Full KV & State KV & Factor & Full ms & State ms\\
\midrule
Qwen3-8B & Addition & 333.54 & 8.16 & 40.89$\times$ & 352.9 & 147.9\\
Qwen3-8B & Modulo 10 & 336.23 & 8.44 & 39.85$\times$ & 382.1 & 75.3\\
Qwen2.5-Coder-14B & Addition & 447.91 & 14.06 & 31.85$\times$ & 533.8 & 154.9\\
Qwen2.5-Coder-14B & Modulo 10 & 451.50 & 14.44 & 31.27$\times$ & 467.4 & 78.6\\
Mistral-7B & Addition & 296.35 & 7.00 & 42.34$\times$ & 1592.6 & 813.0\\
Mistral-7B & Modulo 10 & 298.88 & 7.38 & 40.52$\times$ & 1842.4 & 764.5\\
\bottomrule
\end{tabular}
\end{table}
On these 128-step inputs, state-only prompts require approximately $2.4\%$--$3.2\%$ of the full-history KV payload, a $31$--$42\times$ compression factor. Table~\ref{tab:pilot} reports both accuracy gains and regressions rather than selectively favorable conditions. The KV entries describe numerical tensor payload in a forward pass, not complete GPU process memory, which also includes weights, activations, and allocator reservations. Times are run medians, not cross-machine, hardware-normalized throughput.

\section{Further Implications and Research Vision}
\label{app:scope}
The explicit Markovian structure is a purposeful specialization. Numbering conventions, long-range references, proof branches, and global visited sets need to be included in the current state whenever they are task-relevant; otherwise they cannot be recovered after history is discarded. State-plus-bounded-summary hybrids, occasional full-context anchors, and selected global bypass heads extend coverage where strict compactness is not the right abstraction.

SSMs place compression in a recurrent architecture; our proposal additionally locates its justification in the task dependency graph and the measurable sufficiency of a reduced state. This suggests a tightly integrated outer SSM state evolution and inner Transformer depth recurrence. A general-purpose model can plan, retrieve, and recognize Markovian subproblems; a specialized state kernel can execute repeated structured transitions; the general model can then integrate the result. Such cooperation preserves generality at the system level while specializing the high-frequency execution path.

Future research priorities include adaptive memory-token capacity, native recurrent GPU kernels, process-reward and reinforcement-learning post-training, minimax and information-theoretic bounds under state-sufficiency assumptions, multi-task transition kernels with task-conditioned memory initialization, higher-order and partially observable Markov extensions, and verified interfaces with external calculators, solvers, and simulators. The larger aim is to make long-horizon reasoning a sequence of \emph{state transitions and reusable computation}, rather than an ever-growing transcript.

\section{Notation and Source Attribution}
\label{app:notation}
\begin{table}[h]
\centering
\caption{Core notation.}
\small
\begin{tabular}{ll}
\toprule
Symbol & Meaning\\
\midrule
$Q$ & Fixed task condition / retained prompt\\
$q_t,s_t$ & Reduced state and action / reasoning segment at step $t$\\
$z_t,M_t$ & Compressed state, memory tokens, or recurrent tensor\\
$T,k$ & Outer reasoning steps / chain horizon\\
$R_t,R_{\max}$ & Selected and maximum inner-depth iterations\\
$m,d,d_s$ & Memory-token count, hidden width, and state size\\
$P,s,L_t,L_{\max}$ & Static length, sinks, local length, local budget\\
$N,H,D,b$ & Layers, KV heads, head dimension, bytes per scalar\\
$\lambda_i,\epsilon$ & Training weights and state approximation error\\
\bottomrule
\end{tabular}
\end{table}
This English manuscript translates and reorganizes the Chinese research manuscript by Haibin Tong, Liangliang Xin, Yamei Zhu, and Jiang Yu, preserving its motivation, three-part technical roadmap, design mechanisms, implementation stages, and long-term vision. The three-backbone controlled prompt probe is supplementary and distinctly labeled.

\end{document}